\documentclass{vgtc}                          

\ifpdf
  \pdfoutput=1\relax                   
  \usepackage{graphicx}                
  \DeclareGraphicsExtensions{.pdf,.png,.jpg,.jpeg} 
\else
  \usepackage{graphicx}                
  \DeclareGraphicsExtensions{.eps}     
\fi%

\graphicspath{{figures/}{pictures/}{images/}{./}} 

\usepackage{microtype}                 
\PassOptionsToPackage{warn}{textcomp}  
\usepackage{textcomp}                  
\usepackage{mathptmx}                  
\usepackage{times}                     
\usepackage{cite}                      
\usepackage{tabu}                      
\usepackage{booktabs} 

\usepackage{graphicx}
\usepackage{adjustbox}
\usepackage{makecell}
\usepackage{changepage}

\usepackage{multirow}
\usepackage{tabularx}
\usepackage{caption}
\usepackage{amsmath}

\newcommand{\ie}{\textit{i}.\textit{e}., }

\onlineid{0}

\vgtccategory{Research}

\vgtcinsertpkg

\title{Self-Avatar Animation in Virtual Reality: Impact of Motion Signals Artifacts on the Full-Body Pose Reconstruction}

\author{Antoine Maiorca\thanks{e-mail: antoine.maiorca@umons.ac.be}\\ %
         \scriptsize ISIA Lab, UMONS, Belgium %
\and Seyed Abolfazl Ghasemzadeh \thanks{e-mail: seyed.ghasemzadeh@uclouvain.be}\\ %
  \scriptsize ICTEAM/ELEN, UCLouvain, Louvain-la-Neuve, Belgium %
\and Thierry Ravet \thanks{e-mail: thierry.ravet@umons.ac.be}\\ %
  \scriptsize ISIA Lab, UMONS, Mons, Belgium %
\and François Cresson \thanks{e-mail: francois.cresson@umons.ac.be}\\ %
  \scriptsize ISIA Lab, UMONS, Mons, Belgium %
\and Thierry Dutoit\thanks{e-mail: thierry.dutoit@umons.ac.be}\\ %
  \scriptsize ISIA Lab, University of Mons, Belgium %
\and Christophe De Vleeschouwer\thanks{e-mail: christophe.devleeschouwer@uclouvain.be}\\ %
  \parbox{1.4in}{\scriptsize \centering ICTEAM/ELEN, UCLouvain, Louvain-la-Neuve, Belgium}}

\teaser{
  \centering
  \includegraphics[width=\linewidth]{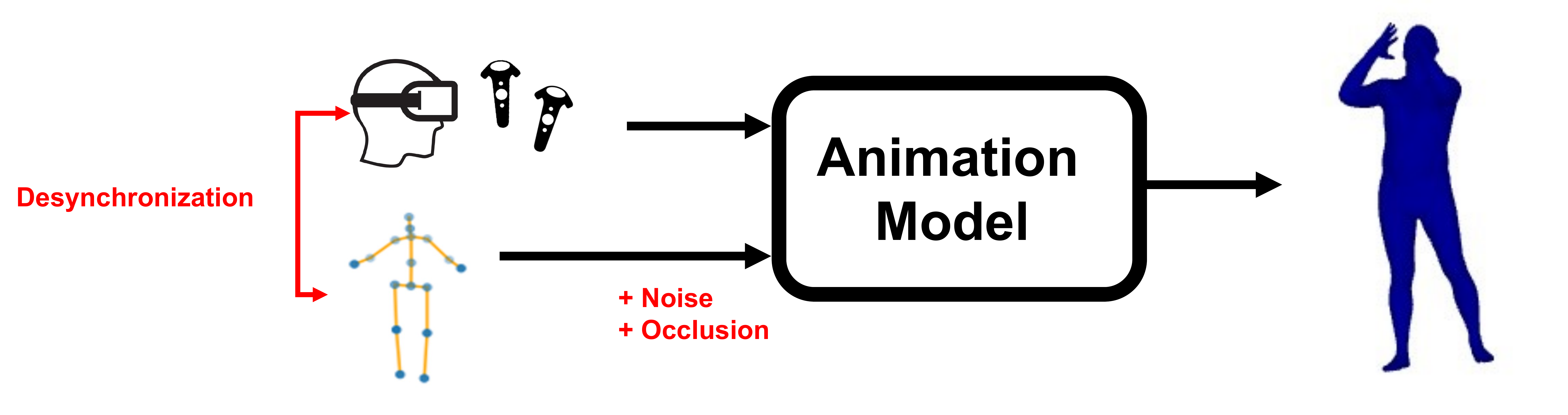}
  \caption{Motion data from sparse inputs are extended with 3D Cartesian positions for leveraging the pose ambiguity in order to reconstruct the full pose of the self-avatar. However, depending on the 3D tracking solution, desynchronization between VR devices motion signals and 3D Cartesian coordinates as well as motion artifacts such as noise and occlusions can arise. With this setup, we can measure the impact of each issue individually in the full-body pose reconstruction.}
  \label{fig:teaser}
}

\abstract{Virtual Reality (VR) applications have revolutionized user experiences by immersing individuals in interactive 3D environments. These environments find applications in numerous fields, including healthcare, education, or architecture. A significant aspect of VR is the inclusion of self-avatars, representing users within the virtual world, which enhances interaction and embodiment. However, generating lifelike full-body self-avatar animations remains challenging, particularly in consumer-grade VR systems, where lower-body tracking is often absent. 
One method to tackle this problem is by providing an external source of motion information that includes lower body information such as full Cartesian positions estimated from RGB(D) cameras. 
Nevertheless, the limitations of these systems are multiples: the desynchronization between the two motion sources and occlusions 
are examples of significant issues that hinder the implementations of such systems. In this paper, we aim to measure the impact on the reconstruction of the articulated self-avatar's full-body pose of (1) the latency between the VR motion features and estimated positions, (2) the data acquisition rate, (3) occlusions, and (4) the inaccuracy of the position estimation algorithm. In addition, we analyze the motion reconstruction errors using ground truth and 3D Cartesian coordinates estimated from \textit{YOLOv8} pose estimation. These analyzes show that the studied methods are significantly sensitive to any degradation tested, especially regarding the velocity reconstruction error.
} 

\CCScatlist{
  \CCScatTwelve{Computing methodologies}{Computer graphics}{Animation};
  \CCScatTwelve{Computing methodologies}{Computer graphics}{Animation}{Motion capture}
}

\begin{document}




\maketitle

\section{Introduction} 

Virtual Reality (VR) applications create a 3D environment that immerses users and enables them to interact with virtual elements. The versatility and interactivity of VR technology have led to a wide array of practical uses spanning various industries and sectors. Besides gaming and entertainment, VR finds valuable applications in healthcare \cite{pillai2019impact, bracq2019virtual}, education and training \cite{jensen2018review, smutny2019review, checa2020review}, as well as architecture and construction engineering \cite{dorta2016hyve, bashabsheh2019application}.

Several studies have highlighted the significance of including a self-avatar, representing the user's body within the virtual environment. These investigations have demonstrated its favorable effects across multiple domains, including user interaction and embodiment \cite{10.3389/frobt.2019.00104}, cognitive processes of participants \cite{7504689}, and collaborative tasks within shared virtual spaces \cite{10.1371/journal.pone.0189078}. Furthermore, the lifelike movements of the avatar contribute to fostering more realistic and engaging social interactions among users within immersive environments  \cite{10.3389/frvir.2021.750729}. Hence, one crucial task in the design of VR applications is the animation of the articulated full-body self-avatar. 

Typically, VR systems designed for consumers consist of a Head-Mounted Display (HMD) along with optional handheld controllers. The HMD is a wearable device worn on the head and positioned in front of the user's eyes to present a visual display. These devices incorporate tracking technology to determine their position and orientation. Using this set of motion data (referred as \textit{sparse inputs} in this document), the system generates the user's complete body movements. However, tracking for the lower body is not included, making it a complex task to synthesize the motion of this body subset. Indeed, synthesizing the full-body motion exclusively from the sparse inputs corresponds to a one-to-many problem since several poses can resolve the motion generation task from one input configuration.

Deep Learning-based animation models have been proposed to tackle this pose ambiguity by learning the complex relationship between the sparse inputs and the lower-body motion \cite{jiang2022avatarposer,zhang2023dual,dittadi2021full,shin2023utilizing}. One significant advantage of these methods is their independence from additional tracking devices, beyond the HMD and controllers, to reconstruct the complete self-avatar's body pose. This feature makes them suitable for consumer-grade applications but the precision in the pose reconstruction can be affected by the lack of available information.

Another approach to mitigate the issue of pose ambiguity involves incorporating external sources of motion data, particularly for the lower-body pose. These additional motion features, that often integrating lower-body limbs information, aim to guide the full-body pose reconstruction \cite{huang2018deep,Kim_Sensors_2022_FusionPoser,WU2019100303,yi2023egolocate}. However, these methods are associated with several drawbacks. Firstly, the need for external devices may hinder the widespread adoption of such technologies. Additionally, equipping users with these sensors can potentially affect comfort and detrimentally impact the overall user experience. To avoid this effect, the full-body Cartesian position can be estimated from RGB videos \cite{8765346,Jocher_YOLO_by_Ultralytics_2023} and act as the additional motion information. 

In this configuration, the latency between VR motion signals and the Cartesian position is a crucial factor that needs to be taken into account in the implementation of a VR animation system. Hence, the process of integrating, into a unified framework, motion capture data from diverse sources, each operating at its distinct framerate, can be a challenging task. Indeed, each sources needs to be synchronized to avoid discrepancies and ensure the coherent representation of the user's movements. This synchronization is essential because misalignments or desynchronization among the data streams can lead to inaccuracies in the reconstructed body pose, which can significantly impact the overall quality and realism of the animation. Moreover, depending of the additional tracking solution and the number of users to track, it can significantly increase the latency between the two motion sources and have implications for the real-time responsiveness of the animation system, which is a crucial factor in this context.

Finally, the pose estimation from RGB(D) videos suffers from a low-fidelity pose reconstruction compared to physical motion capture devices such as optical markers. Low accuracy on the joint position and artifacts such as occlusion \ie a hidden information resulting in a missing joint in the estimated pose, are major concerns for animating self-avatar's based on this set of motion features.

The goal of this paper is to analyze the impact in the articulated self-avatar's full-body pose reconstruction of:
\begin{itemize}
    \item the latency between the Cartesian position estimated from RGB videos and the VR motion signals
    \item the discrepancy between the motion acquisition rate of these two sources of motion
    \item the motion artifacts that can occur in the estimation of the full-body Cartesian coordinates
\end{itemize}

To do so, we propose to manually degrades the 3D Cartesian positions with the artifacts described as above. The animation model is then fed by the sparse inputs concatenated with these 3D positions. Finally, we assess the reconstruction error across the tested configurations. Moreover, we suggest extracting the 3D Cartesian coordinates using the pose estimation algorithm embedded in \textit{YOLOv8} \cite{Jocher_YOLO_by_Ultralytics_2023}. This approach enables us to discern the disparities in model performance when utilizing either ground truth or 3D Cartesian positions estimated in a real-world use case.

\section{Related Work}
Approaches addressing the animation of a full-body self-avatar in VR can be broadly categorized into two main groups: those relying solely on sparse inputs and those augmenting VR motion signals with extra motion data. These approaches are respectively explained in Section \ref{subsec:rw1} and \ref{subsec:rw2}.    


\subsection{Self-avatar's full-body estimation from sparse VR sensors}
\label{subsec:rw1}
Tracking a user's full-body motion based on sparse sensors is a widely explored topic \cite{Kim_Sensors_2022_FusionPoser,kim2012realtime,Yi_2022_CVPR,s20216330}. In the VR paradigm, the full motion tracking is performed using the motion signals from the VR devices. The self-avatar's full-body motion is synthesized from the hands and head motion features. Solutions based on Inverse Kinematics (IK) have been implemented to leverage this problem \cite{lang2016inverse,jiang2016real,tan2017virtual,inproceedings}.

Then, machine learning techniques have further been used to improve the quality of the full-body motion reconstruction. In the case of \textit{CoolMoves} \cite{ahuja2021coolmoves}, it leverages k-NN techniques to find a pose that closely corresponds to the sparse inputs within a well-structured motion database. Similarly, \textit{MMVR} \cite{ponton2022combining} adopts motion matching \cite{buttner2015motion} as an alternative approach to achieve real-time animation with smooth transitions. The major drawback of these approaches is the the fact that we must ensure that the motion database gathers realistic samples that include not only seamless transitions and smooth blending between distinct motions but also a diverse array of desired actions. Moreover, animating upper body gestures using motion matching is challenging due to the unconstrained nature of user's arm movements. This complexity necessitates the creation of an extensive and challenging-to-manage motion database to encompass the multitude of feasible poses.

With the emergence of Deep Learning, methods based on artificial neural networks have been designed to capture the complex spatial-temporal dependencies between the sparse signals and the full body pose, especially regarding the low body information where no tracked data are available. More specifically, algorithms built upon architectures designed for time series analysis have been proposed to compute the self-avatar's full pose based on the sparse inputs \cite{jiang2022avatarposer,yang2021lobstr,zhang2023dual,shin2023utilizing}. While LSTMs have been successfully employed in this context \cite{yang2021lobstr}, Transformer-based methods outperformed this solution regarding the quality of the pose reconstruction. \textit{AvatarPoser} \cite{jiang2022avatarposer} has integrated the encoder part of the transformer architecture \cite{vaswani2017attention} to compute high-dimensional embeddings from the sparse information to estimate the avatar's full pose and its global displacement. \textit{Dual Attention Poser} \cite{zhang2023dual} have proposed to decouple the global and local motion features to feed two transformers and further merge the computed information. 

An alternative approach to address this challenge involves generative models. \textit{VAE-HMD} \cite{dittadi2021full} trained a Variational AutoEncoder (VAE) to generate a latent vector from a sequence of sparse inputs. This latent vector is then used as input to the decoder, which is responsible for reconstructing the complete body pose. More recently, methods based on the combination of transformers and generative algorithms have been introduced: In the same philosophy to \textit{VAE-HMD}, full motion prior is used as a part of a pretraining process and  a transformer-based encoder is then trained to predict the same motion latent as the full motion encoder using sparse inputs \cite{shin2023utilizing}. Additionally, \textit{FLAG} \cite{aliakbarian2022flag} employs the technology of Normalizing Flow \cite{rezende2015variational} that allows to compute \textit{exact} pose likelihoods in contrast with VAEs and improved the quality of the full pose reconstruction in comparison with \textit{VAE-HMD} outputs. \textit{BoDiffusion} \cite{castillo2023bodiffusion} employs a generative diffusion model \cite{sohl2015deep} for motion synthesis to tackle this under-constrained reconstruction problem.

Finally, the full pose reconstruction from sparse inputs problem has been leveraged involving physical-simulation of the self-avatar's body \cite{winkler2022questsim,lee2023questenvsim}. These methods are built upon Reinforcement Learning for a physically plausible full-body pose reconstruction.  

\begin{figure*}
    \centering
    \includegraphics[width=0.7\textwidth]{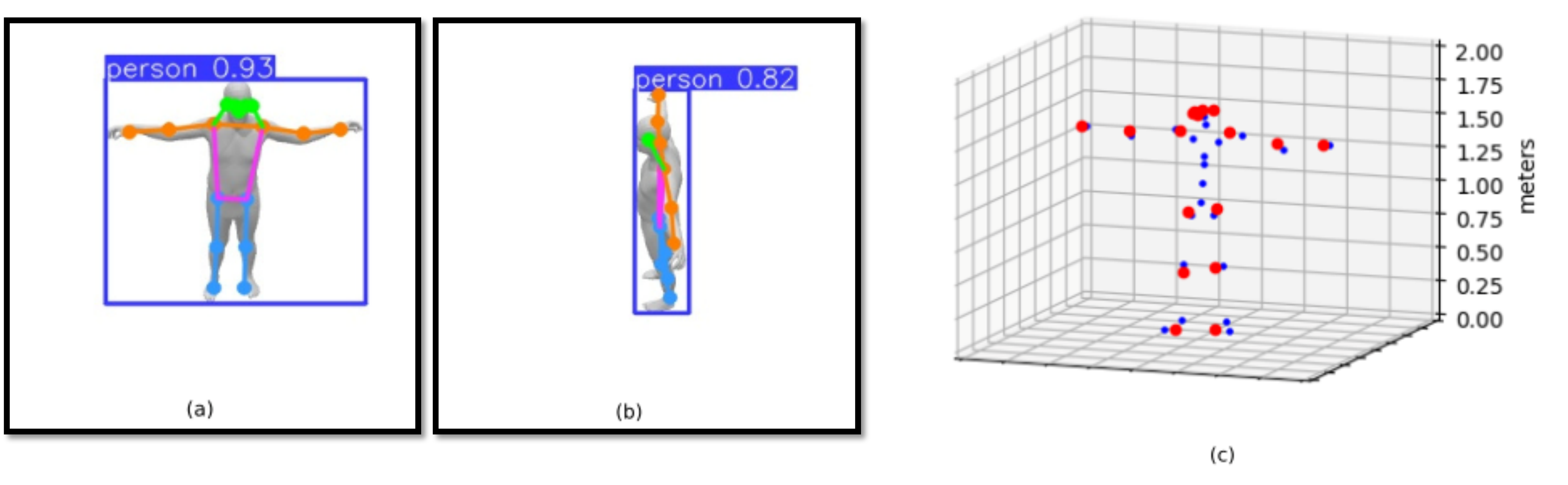}
    \caption{Reconstruction of the full-body position  based on YOLOv8 algorithm. (a) and (b) Illustration of YOLOv8 detection applied to one frame of AMASS dataset rendered from two different perspectives. (c) the 3D Cartesian reconstruction pose computed by triangulation (red) compared with ground truth frame (blue)}
    \label{fig:yoloReconstruction}
\end{figure*}

\subsection{Multimodal full-body pose estimation}
\label{subsec:rw2}

In order to resolve the pose ambiguity, external motion sources have been deployed to extend the information tracked by the HMD and the handheld controllers. 

Inspired by the success of 2D pose estimation from monocular RGB videos, \textit{HybridTrack} \cite{yang2022hybridtrak} extends VR motion data with the 2D Cartesian coordinates. By combining an uncalibrated RGB camera for the lower body with inside-out tracking for the upper body, it offers cost-effective and user-friendly tracking capabilities. 

Moreover, Liao et al. in \cite{reconstruction3d:2023:PeerJ} explore the fusion of sparse Inertial Measurement Units (IMUs) with a single RGB camera to achieve robust 3D human body reconstruction without the need for invasive multi-IMU setups or 2D joint detection. Using adaptive regression learning, a dual-stream network extracts features from IMUs and images, followed by a residual model-attention network that effectively fuses these features. The method significantly improves upon existing approaches by reducing errors in 3D joint positions and enhancing robustness, particularly in scenarios with occlusions or challenging environments.

Next, \textit{EgoLocate} \cite{yi2023egolocate} emphasizes the importance of merging human motion sensing and environment sensing. They note that while human motion capture and environment sensing, such as SLAM, are crucial, they have traditionally been treated as separate domains. The authors introduce their \textit{EgoLocate} system, which integrates sparse IMUs and a monocular camera to achieve real-time human motion capture, localization, and mapping. 

Furthermore, Tome et al. in \cite{selfpose:2023:tome} propose a novel neural network architecture to address challenges such as strong perspective distortions, self-occlusion, a lack of labeled datasets, and the inherent ambiguity in lifting 2D joint positions to 3D space, demonstrating significant improvements in performance compared to existing methods, both in egocentric and front-facing camera scenarios.

Finally, \textit{BodyTrak} \cite{bodyTrack:2022} employs wrist-mounted cameras to capture informative images of body silhouettes and derives a working hypothesis that these partial body parts can effectively infer full-body poses. The contributions include the introduction of the first wrist-mounted device for full-body pose estimation, a custom deep learning pipeline, user studies, and discussions on the challenges and opportunities of applying such technology in practical contexts.


\section{Problem Formulation}
\label{sec:prob}

\begin{figure*}[!htb]
    \centering
    \includegraphics[width=\textwidth]{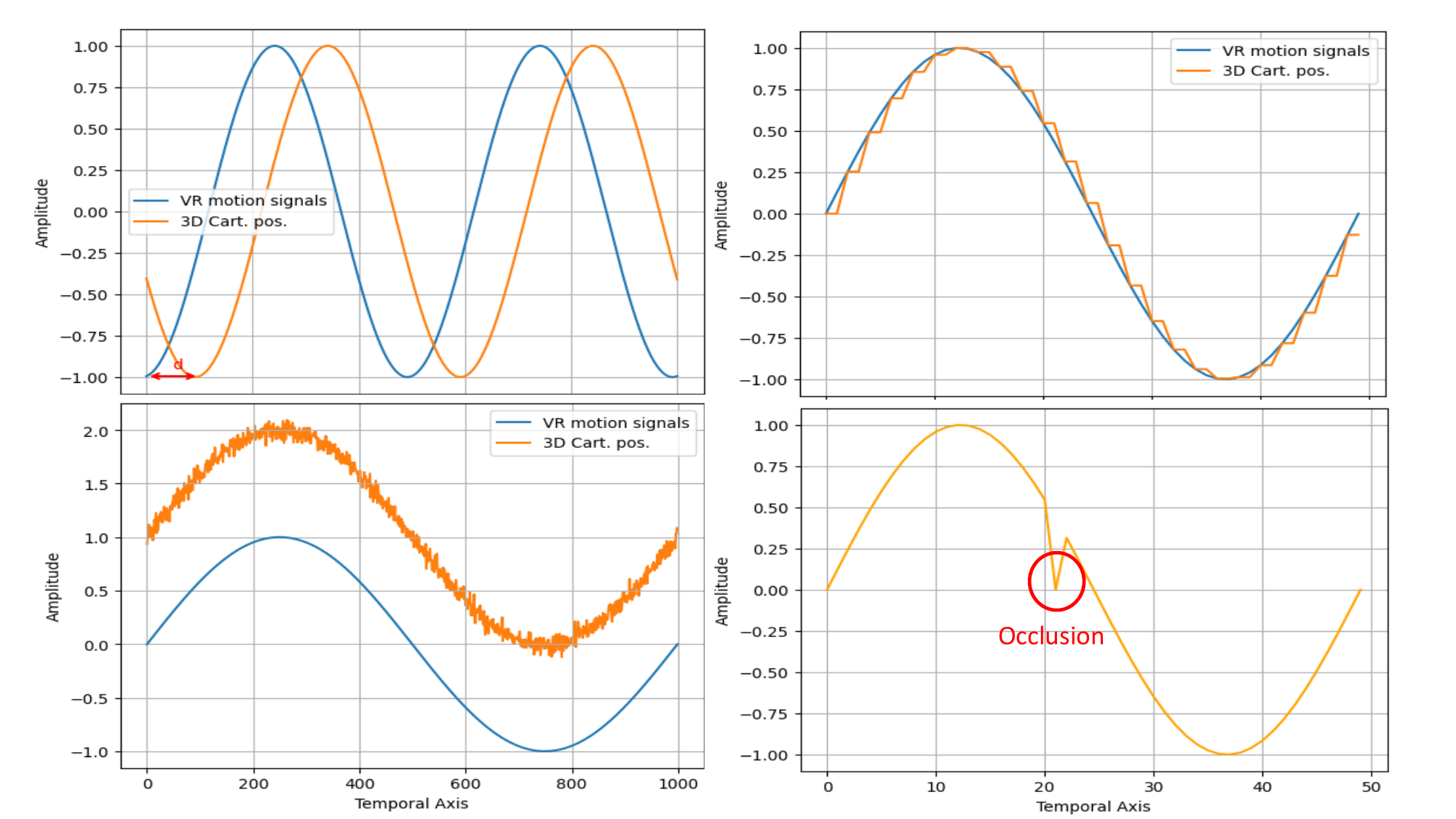}
    \caption{Examples of artifacts on motion signals. \textbf{Top left:} delay between two motion sources. \textbf{Top Right:} Cartesian position framerate reduced by $fps_{ratio} = 2$. \textbf{Bottom Left:} Gaussian noise applied on positional data. \textbf{Bottom Right:} Random occlusion \ie a joint position randomly set to zero.}
    \label{fig:degr}
\end{figure*}

The generation of the self-avatar's full-body pose can be formalized as follow: considering a sequence of the sparse inputs $X_{0,...,T}$ and additional motion data $X^F_{0,...,T}$, a method $\Phi$ is designed to synthesize the full pose of the articulated self-avatar $Y$, as in Equation \ref{eq:method}.

\begin{equation}
    \label{eq:method}
    Y_T = \Phi(X_{0,...,T} \oplus X^F_{0,...,T})
\end{equation}

Similarly to \textit{AvatarPoser} \cite{jiang2022avatarposer}, $X_{0,...,T}$ gathers the Cartesian positions, the orientations, the linear and angular velocities of the head and the hands. We choose to extend the sparse inputs with the 3D Cartesian positions that can be estimated with a setup of one or multiple RGB(D) cameras \cite{Hartley:2003:MVG:861369} This configuration has the advantage to not require any physical cumbersome tracking sensors on the user body which can infringe with its overall experience.

However, implementing this kind of solution is not a straightforward task since it deals with several constraints. In this context, the most common concerns are the lack of synchronization between the two motion sources, induced by a latency and a discrepancy between the data acquisition rate from both sources, joints occlusion and the low-fidelity of the Cartesian position estimation method. 

\begin{itemize}
    \item \textbf{Latency}: Depending on the cameras setup, a significant latency between the Cartesian positions and the sparse inputs can occur. Indeed, the head and hands positions and orientations are computed from inertial sensors or using SLAM-based algorithms \cite{jinyu2019survey} regarding the VR devices employed. The delay introduced by this process is relatively minor, typically ranging from 1 to 2 milliseconds when employing IMU-based head-mounted displays operating at $1000Hz$ for the rotational data and between $60Hz$ and $100Hz$ for the positional data. In contrast, the overall position estimation system often introduces a more substantial delay in the context of the deployed solution. Moreover, this latency is notably variable over time, primarily owing to frame loss. For instance, when assessing the \textit{ZED 2} stereo camera \cite{zed2} operating at a 60 FPS configuration, it exhibits a latency range of 30-45ms. This leads to desynchronization between the head-mounted display and the camera frames.
    \item \textbf{Framerate}: The motion data acquisition rates are not necessarily equivalents between the motion sources. The motion features of the HMD and the hands are extracted at $60Hz$ or $100Hz$ regarding the VR devices used \cite{vr}. While some cameras, such as the \textit{ZED 2}, can also achieve frame rates of 60 or even 100 FPS, it's crucial to take into account the computational demands associated with 3D position estimation\cite{Jocher_YOLO_by_Ultralytics_2023}. This computational load can become substantial, contingent on the chosen algorithm and the number of individuals being tracked within the scene.
    \item \textbf{Occlusion}: It refers to the situation where a user being tracked is partially or completely hidden from the cameras used for tracking. This obstruction can occur when one user passes in front of another, or when a part of the user's body blocks the view of one of its limbs. Occlusion poses a challenge for motion tracking systems because it can lose sight of one or multiple joints during the occluded period, leading to gaps or inaccuracies in the tracking pose. In our configuration, since we consider the additional motion information $X^F$ as the 3D Cartesian positions estimated from cameras, our setup is subject to occlusion artifacts. This artifact gains prominence, particularly in scenarios involving multiple users interacting with each other, thereby increasing the likelihood of occluded joints.
    \item \textbf{Low accuracy}: Since the markerless motion tracking algorithm based on computer vision are often less accurate and reliable than optical markers, the low fidelity on the 3D Cartesian coordinates estimation can have a negative impact on the full-body pose reconstruction.  
\end{itemize}






\section{Approach}

As in \textit{AvatarPoser} \cite{jiang2022avatarposer}, we rely on 3 subsets of the large-scale motion capture AMASS Dataset \cite{AMASS:ICCV:2019}: BMLrub \cite{troje2002decomposing}, CMU \cite{cmu} and HDM05 \cite{muller2007documentation}. AMASS Dataset unifies optical-based motion capture datasets into a standard kinematic tree and use SMPL approach \cite{loper2015smpl} to provide realistic 3D human meshes represented by a rigged body model. These subsets gather around 5200 motion samples for a duration of more than 20 hours. The standardized kinematic tree is structured into 22 joints.  

In order to measure the impact of each artifacts individually, we propose to mimic the artifacts described in Section \ref{sec:prob}. Examples of these artifacts on motion signals are shown in Figure \ref{fig:degr}.


\begin{itemize} 
    \item We introduce a delay $d$ between the full-body Cartesian positions and the sparse motion signals. In this configuration, Equation \ref{eq:method} becomes 
        \begin{equation}
            Y_{T+d} = \Phi(X_{d,...,T+d} \oplus X^F_{0,...,T})
        \end{equation}
    \item The framerate discrepancy between the two motion signals is tackled by quantifying $X^F_{0,...,T}$ using a framerate ratio $fps_{ratio}$ between the two motion sources. The framerate of $X^F_{0,...,T}$, initially equivalent to $X_{0,...,T}$, is divided by $fps_{ratio}$.
    \item Inspired by the implementation of occlusion in \cite{holden2018robust}, we define a probability $\sigma_o$ of the joint being occluded for a given frame.
        \begin{align}
            \label{eq:occl}
            O^{F}_{0,...,T} \sim Bernoulli(\sigma_o) \\
            X^{F}_{o_{0,...,T}} = X^{F}_{0,...,T} * O^{F}_{0,...,T} \\
            Y_{o_T} = \Phi(X_{0,...,T} \oplus X^F_{o_{0,...,T}})
        \end{align}

    \item To mimic the low accuracy of the full-body position estimation, we add noise sampled from a zero-mean Gaussian distribution with a standard deviation $\sigma$. Increasing the standard deviation $\sigma$ results in an augmented level of noise intensity within the Cartesian positions. 
        \begin{align}
            N_{0,...,T} \sim \mathcal{N}(0, \sigma^2) \\
            X^{F}_{n_{0,...,T}} = X^{F}_{0,...,T} + N_{0,...,T} \\
            Y_{n_T} = \Phi(X_{0,...,T} \oplus X^F_{n_{0,...,T}})            
        \end{align}

    However, this distribution may not accurately represent the actual noise in the markerless pose estimation system. To build a set of noisy poses that reflects the lack of accuracy of the data we could acquire in a real-case scenario, we use a solution based on the YOLOv8 algorithm \cite{Jocher_YOLO_by_Ultralytics_2023}. As shown in Figure \ref{fig:yoloReconstruction}, using the SMPL+H model \cite{MANO:SIGGRAPHASIA:2017} linked to AMASS dataset, we define two virtual cameras to render the character as mesh from two different viewpoints. The rendering resolution in pixels is 640X480. The processing of each rendered image with Yolo v8 detector (yolov8x-pose model) provides a pose as a set of characteristic points. We apply the triangulation function from the OpenCV library \cite{opencv_library} to the characteristic points in the two perspectives, using the parameters of the virtual cameras as explained in \cite{Hartley:2003:MVG:861369}. The result is a 3-dimensional reconstruction of the pose computed for each frame in the dataset. 
\end{itemize}

\section{Experiments}

\begin{table*}[htb]

\caption{Evaluation of the sensitivity to common artifacts in the context of the self-avatar's animation from sparse inputs and 3D Cartesian coordinates. The highlighted results refer to configurations that outperform \textit{AvatarPoser} with sparse inputs. We observe that \textit{AvatarPoser} overall gives the best results when it comes to deal with clean motion data. However, as a matter of example, when the noise or occlusion intensity increases, \textit{HybridTrack} leads to lower reconstruction error.} 
\centering


\begin{tabularx}{\textwidth}{|X|X||X|X|X|X|X|X|}

\hline
 & Body subset & \multicolumn{3}{c|}{\textit{AvatarPoser}} & \multicolumn{3}{c|}{\textit{HybridTrack}} \\
\cline{3-8}
& & MPJPE (cm) $\downarrow$ & MPJRE (°) $\downarrow$ & MPJVE (cm/s) $\downarrow$ & MPJPE (cm)$\downarrow$ & MPJRE (°)$\downarrow$ & MPJVE (cm/s)$\downarrow$ \\
\hline
Sparse inputs & Up & 1.65 & 5.64 & 12.86 & - & - & - \\
Sparse inputs & Low & 6.79 & 6.4 & 44.35 & - & - & - \\
\hline
GT 3D Cart. Pos. & Up & \textbf{0.72} & \textbf{2.52} & \textbf{8.1} & 2.38 & 6.51 & 20.72 \\
GT 3D Cart. Pos. & Low & \textbf{1.41} & \textbf{2.03} & \textbf{14.19}& \textbf{4.97} & \textbf{5.55} & \textbf{40.81} \\
\hline
Noise $\sigma (cm)$ & & & & & & &  \\
\hline
1 & Up & \textbf{1.09} & \textbf{3.89} & 60.78 & 2.39 & 6.58 & 35.39 \\
1 & Low & \textbf{1.96} & \textbf{3.37} & 107.91 &\textbf{ 5.02} & \textbf{5.61} & 69.86 \\
\hline
2 & Up & 1.74 & 6.33 & 121.98 & 2.46&	6.78	&55.94 \\
2 & Low &  \textbf{3.04}	&\textbf{5.65}	&215.17 & \textbf{5.18	}&\textbf{5.77	}&111.11\\
\hline
5 & Up & 4.18 & 15.04 & 326.92 & 2.92&	8.13	&118.89\\
5 & Low & 7.2&	13.35	&565.61 & \textbf{6.25	}&6.83	&240.02\\
\hline
Occlusion $\sigma_o$ & & & & & & & \\
\hline
0.01 & Up & 3.05&	8.18	&272.48& 4.05	&10.16	&231.65\\
0.01 & Low &\textbf{ 5.87}	&\textbf{5.26}	&525.83&  7.87	&7.59	&414.22\\
\hline
0.05 & Up& 10.36	&23.47	&912.65 & 8.95	&19.14	&525.12\\
0.05 & Low &19.68	&13.73	&1756.09& 15.82	&13.13	&992.5\\
\hline
Framerate ratio $fps_{ratio}$ & & & & & & & \\
\hline
2 & Up & \textbf{0.87}	&\textbf{2.73}	&23.98 & 2.43	&6.54	&21.96 \\
2 & Low & \textbf{1.81}	&\textbf{2.41}	&57.02& \textbf{5.19}	&\textbf{5.71}	&45.29 \\
\hline
3 & Up & \textbf{1.05}	&\textbf{3.03}	&30.91&2.49	&6.6	&27.71\\
3 & Low &\textbf{2.22}	&\textbf{2.81}	&74.18& \textbf{5.44}	&\textbf{5.89}	&68.37\\
\hline
4 & Up & \textbf{1.22}	&\textbf{3.35}	&34.47& 2.55	&6.66	&25.88\\
4 & Low & \textbf{2.64}	&\textbf{3.21}	&82.65& \textbf{5.72}	&\textbf{6.1}	&57.54\\
\hline
Delay $d$ (frames)  & & & & & & & \\
\hline
2 & Up & \textbf{1.44}&	\textbf{3.78}	&14.36& 2.62	&6.73	&21.77\\
2 & Low & \textbf{3.02}	&7.23&	\textbf{24.57}&  \textbf{5.9}8	&\textbf{6.31}	&47.5\\
\hline
4 & Up & 2.24	&\textbf{5.05}	&19.8&2.95	&7.12	&23.28\\
4 & Low &\textbf{4.83}	&\textbf{5.42}	&42.69& 7.26	&7.31	&56.27\\
\hline
6 & Up & 3.02	&7.23	&24.57& 3.34	&7.63	&25.22\\
6 & Low & \textbf{6.5}	&7.01	&54.15& 8.66	&8.36	&65.51\\
\hline
\end{tabularx}

\label{tab:res}
\end{table*}

For our experiments, we train the animation model $\Phi$ with $X_{0,...,T} \oplus X^F_{0,...,T}$ as input where $X^F_{0,...,T}$ is a sequence of the ground truth 3D Cartesian positions and then measure the impact of each artifacts independently. We rely on \textit{AvatarPoser} \cite{jiang2022avatarposer} and an adapted version of \textit{HybridTrack} \cite{yang2022hybridtrak}, two state-of-the-art deep learning-based models for the animation of the self-avatar's full-body. \textit{AvatarPoser} \cite{jiang2022avatarposer} is a model built upon the Transformer architecture that encodes the sparse inputs into a high-level complex motion representation to estimate the full-body local orientations as well as the global displacement of the root. Providing 3D Cartesian positions into this model helps to resolve the pose ambiguity that can arise due to the lack of lower body motion information. \textit{HybridTrack} \cite{yang2022hybridtrak} employs a CNN-1D based architecture inspired by the method in \cite{pavllo20193d}, that is fed by the sparse information and a single-view of 2D Cartesian coordinates to generate the full body pose. In our experiments, we adapt this model to our requirements by providing the 3D Cartesian positions instead. 

We add a latency between the Cartesian positions and the sparse inputs of $d = 2$, $4$ or $6$ frames following the delays discussed in Section \ref{sec:prob}. Then, since the framerate is sensitive to the computational load induced by several factors such as the number of cameras or the number of tracked users, we divided the Cartesian coordinates framerate by a factor $fps_{ratio} = 2$, $3$ or $4$. In this last case, if the motion data acquisition rate from VR devices reaches $100Hz$, the cameras achieve a framerate of $25Hz$.
Regarding the spatial-temporal noise, we set up $\sigma_o = 0.01$ and $0.05$ of occlusion probability ($\sigma_o = 0.1$ in \cite{holden2018robust}) and $\sigma = 1cm$, $2cm$ and $5cm$ which are reasonable noise levels considering the error reconstruction in multi-users 3D pose estimation tracking from monocular videos  \cite{zanfir2018deep} ($60mm$ on \textit{Human36M} dataset \cite{h36m_pami}).


Finally, we test our implementations in a real-case scenario: from AMASS Dataset, we use the pose estimation from \textit{YOLOv8} to gathers the 3D Cartesian positions, instead of using the ground truth Cartesian positions to train and to evaluate a specific animation model.

We consider a temporal window of $40$ previous frames and the current frame ($T=41$ frames) feeding the models. Both of them are trained during 10k epochs on a NVIDIA GTX1080 GPU, following the training procedure and hyperparameters in \textit{AvatarPoser}\cite{jiang2022avatarposer} \footnote{\href{https://github.com/eth-siplab/AvatarPoser.git}{https://github.com/eth-siplab/AvatarPoser.git}}. 

\section{Results}

\begin{figure*}[!htb]
    \centering
    \includegraphics[width=0.8\textwidth]{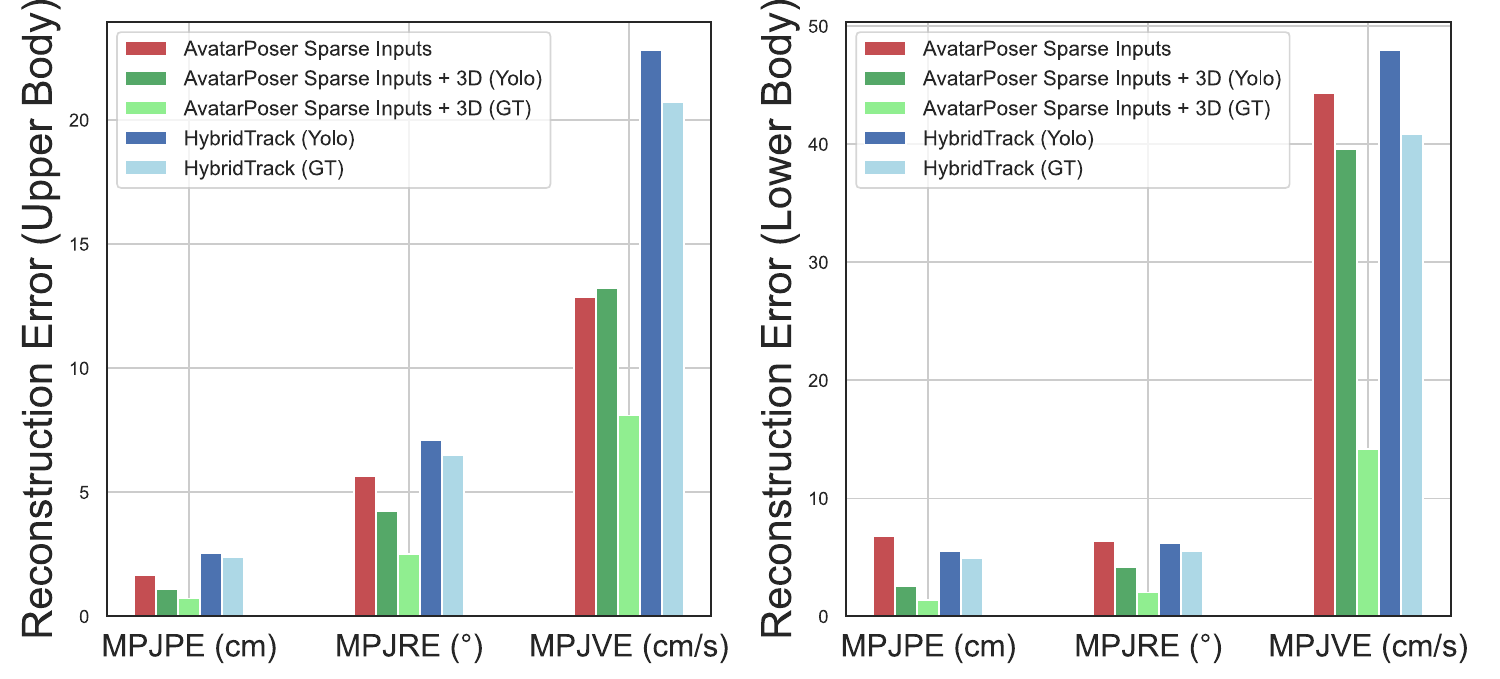}
    \caption{Reconstruction errors regarding the models trained with ground truth and Cartesian positions from YOLOv8.}
    \label{fig:yolo_res}
\end{figure*}

Table \ref{tab:res} shows the results of the evaluation of the two animation models to the considered motion signals artifacts. We rely on 3 metrics called the Mean Per Joint Positional/Rotational/Velocity Error, respectively referred as \textit{MPJPE}, \textit{MPJRE} and \textit{MPJVE}. It indicates the deviation between the avatar's full pose estimated by the model and the ground truth pose. We also refer in the aforementioned Table the results on the full pose reconstruction from $X_{0,...,T}$  for \textit{AvatarPoser} but not for \textit{HybridTrack} since it has not been designed to tackle the problematic of the full pose reconstruction from only the sparse inputs. 

We observe that the overall behavior is that the errors increase with the intensity of the artifact for both models and regardless the evaluated artifact. The models trained with the clean and synchronized data from AMASS dataset does not encompass the issues related to the animation of the self-avatar's character based on the sparse inputs and the Cartesian positions estimated from cameras. Then, when the spatio-temporal motion artifacts, \ie the occlusions and the noise, increases in intensity, \textit{HybridTrack} provides more accurate full pose reconstruction than \textit{AvatarPoser}. In this configuration, \textit{HybridTrack} exhibits greater resilience and accuracy than \textit{AvatarPoser}, showcasing its enhanced performance in challenging conditions marked by heightened motion artifacts. However, it is deemed ineffective to incorporate 3D Cartesian positions in both animation model when confronted with intense artifacts. Indeed, while \textit{HybridTrack} effectively mitigates the impact of these artifacts, \textit{AvatarPoser}, trained solely with $X_{0,...,T}$, outperforms \textit{HybridTrack} across various configurations. We highlighted in Table \ref{tab:res} the configurations where the models fed by $X_{0,...,T} \oplus X^F_{0,...,T}$ exhibits more accurate pose reconstruction than \textit{AvatarPoser} trained with sparse inputs. The tested artifacts appear to have a more pronounced effect on velocity compared to positions and orientations. In summary, this evaluation highlights the sensitivity of the models to variations in the training parameters, limiting their deployment in less controlled environments.

The results, reported in \ref{fig:yolo_res}, show the comparison between models trained with ground truth and models trained with Cartesian positions generated with YOLO. \textit{HybridTrack} produces less accurate motion reconstruction than \textit{AvatarPoser} with solely the sparse inputs regarding the upper body. However, \textit{AvatarPoser} trained with the 3D coordinates estimated from YOLOv8 outperforms \textit{AvatarPoser} from sparse inputs, except for the velocity of the upper body region. Comparing these two cases, we can also observe that we do not get any significant difference in velocity in the lower body region either. So, there is no clear improvement in terms of noise present in the synthesized movement.

\begin{figure}[!htb]
    \centering
    \includegraphics[scale=0.13]{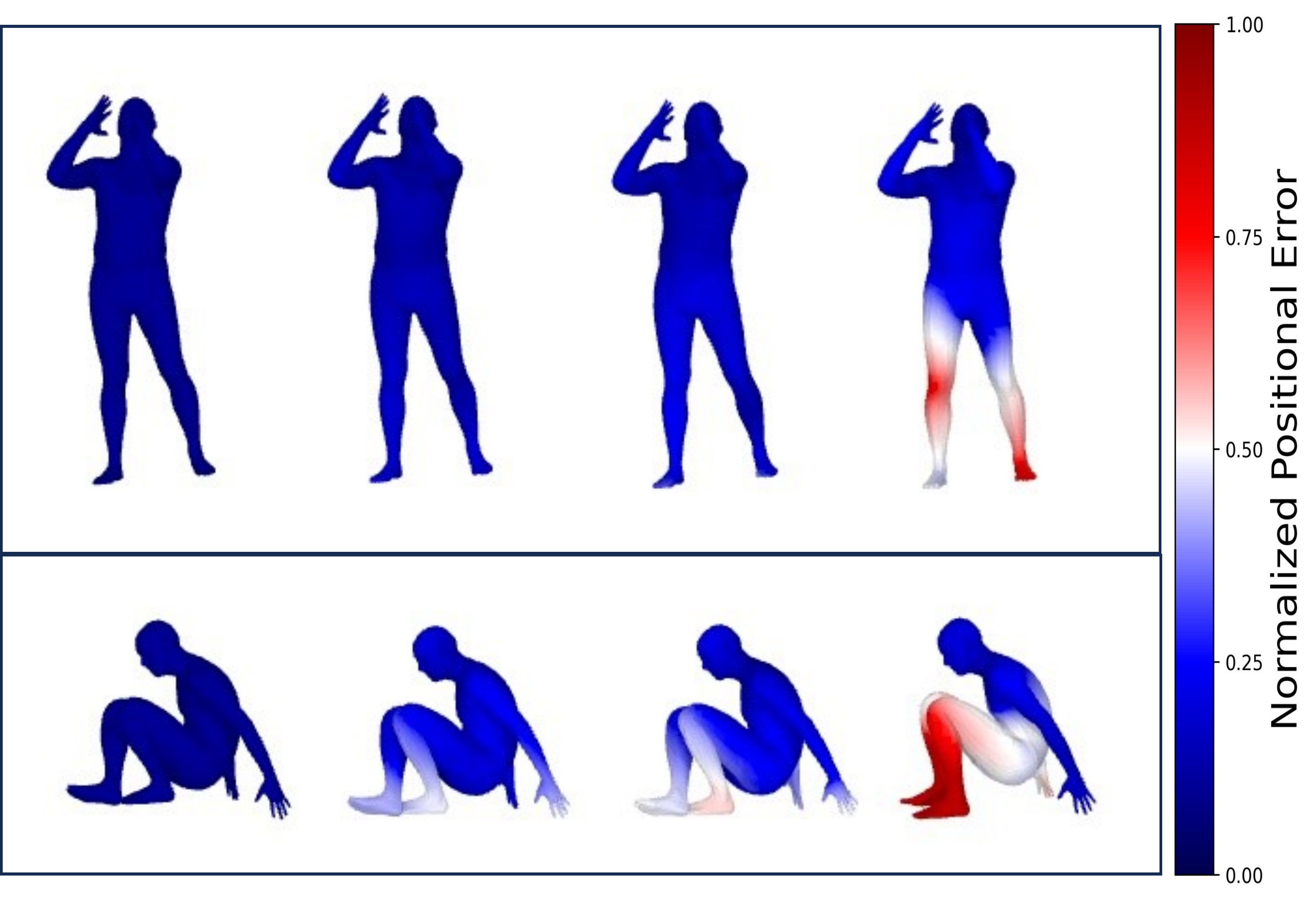}
    \caption{Illustration of two pose samples derived from the ground truth data (\textbf{Left}). In the \textbf{Mid-Left} image, \textit{AvatarPoser} is trained with ground truth 3D Cartesian positions and provided with a sequence of this ground truth. The \textbf{Mid-Right} image displays the effects of introducing Gaussian noise into the 3D Cartesian coordinates, with noise levels $\sigma = 0.01$. \textbf{Right}: \textit{AvatarPoser} trained with only sparse inputs. The positional features are improved in comparison to those produced by solely the sparse inputs, even when the Cartesian coordinates are degraded with a Gaussian noise with $\sigma=1cm$.}
    \label{fig:pose}
\end{figure}

Illustrative pose samples are depicted in Figure \ref{fig:pose}. The coloration on various regions of the mesh represents the positional error specific to each region. The figure serves to highlight that the quality of motion reconstruction diminishes as noise is introduced to the 3D coordinates. More videos presenting motion samples can be found here \footnote{\href{https://figshare.com/s/a6f9c9770e4be4919230}{https://figshare.com/s/a6f9c9770e4be4919230}}

\section{Limitations and Perspectives}

First of all, our analyses emphasize the sensitivity of \textit{AvatarPoser} and \textit{HybridTrack} when trained with ground truth Cartesian positions and sparse inputs from VR devices. While it has been shown that incorporating Cartesian positions aids both models in addressing this challenge, the artifacts examined in this study—such as latency between motion sources, discrepancies in framerate, low accuracy of pose reconstruction, and occlusions—result in a degradation of the quality of the self-avatar's full-body pose reconstruction.



Then, even if we consider these effects in the training process, both of the models failed to improve the upper-body velocities in comparison to the model trained only with the sparse inputs. Mitigating this effect will be crucial for further development, especially when several users share the scene captured by the cameras. Indeed, in this context, the risk of major artifacts such as occlusions can explode.


Finally, considering the desynchronization between the two motion signals, recent methods have been proposed to integrate temporally sparse observations in Transformer within a medical context \cite{tipirneni2022self}. We believe that the integration of such systems will benefit the field of self-avatar's animation regarding the issues discussed in this work.


\section{Conclusion}

In this work, we conducted experiments in the context of the self-avatar's full-body pose reconstruction from the head and hands motion features and the 3D Cartesian coordinates. This additional motion data has been chosen so that it can be extracted from a non-intrusive systems, such as RGB-D cameras, to guarantee the user experience. First, we discussed about the major concerns that need to be taken into account in the design of such system. The desynchronization between the motion signals from the VR devices and the Cartesian positions, as well as the spatio-temporal motion artifacts such as noise and occlusions are artifacts that importantly degrades the reliability of the animation system. We showed that the precision of the pose estimation system is crucial in this context, especially considering the velocity of the reconstructed upper body. From the results of these analyzes, we believe that this work provides valuable insights concerning the task of self-avatar's animation based on multimodal data.

\acknowledgments{
This research is supported and funded by TRAIL Institute. S.A. Ghasemzadeh is funded by FRIA/FNRS.}

\bibliographystyle{abbrv-doi}

\bibliography{template}
\end{document}